\documentclass{article}
\usepackage{bbm}
\usepackage{hhline}
\usepackage{multirow}
\usepackage{array,multirow,graphicx}
\usepackage{arydshln}
\usepackage{amsmath}

\usepackage{times}
\usepackage{xcolor}
\usepackage{booktabs}
\usepackage{longtable}
\usepackage{supertabular}

\title{A Noise-Robust Loss for Unlabeled Entity Problem in Named Entity Recognition}
\author{Wentao Kang, Guijun Zhang, Xiao Fu}

\begin{document}

\maketitle

\begin{abstract}
Named Entity Recognition (NER) is an important task in natural language processing. However, traditional supervised NER requires large-scale annotated datasets. Distantly supervision is proposed to alleviate the massive demand for datasets, but datasets constructed in this way are extremely noisy and have a serious unlabeled entity problem. The cross entropy (CE) loss function is highly sensitive to unlabeled data, leading to severe performance degradation. As an alternative, we propose a new loss function called NRCES to cope with this problem. A sigmoid term is used to mitigate the negative impact of noise. In addition, we balance the convergence and noise tolerance of the model according to samples and the training process. Experiments on synthetic and real-world datasets demonstrate that our approach shows strong robustness in the case of severe unlabeled entity problem, achieving new state-of-the-art on real-world datasets.

\end{abstract}

\section{Introduction}

Named entity recognition (NER) is a critical task in information extraction. It aims to identify meaningful entity mentions in natural languages, such as names of persons, locations, organizations, etc. In recent years, traditional supervised methods for NER have achieved great success due to the neural network's strong representational power (Li et al.,2020). However, their effectiveness mainly relies on large-scale, high-quality datasets, which are expensive and labor-intensive to collect.

To address this problem, distantly supervised named entity recognition has been proposed. An intuitive and common way is aligning entities in dictionaries or knowledge bases to corresponding entity spans in the corpus. It eliminates the cost of human annotations but introduces noisy labels due to the limited coverage of knowledge resources. Weakly annotated data suffers from severe unlabeled entity problem. In other words, many entity mentions are mislabeled as non-entities. If the NER model overfits these wrongly introduced negative samples, it will seriously fail with poor performance.

Several methods are typically used to mitigate the effects of noise. For incompletely annotated data, some works redefined new deep learning paradigms (Yang et al.,2018; Peng et al.,2021; Zhang et al.,2021). Some works address this problem based on negative sampling. Li et al. (2020) firstly defined unlabeled entity problem, and used negative sampling to avoid the misleading of unlabeled entities. In their latest work (Li et al.,2022), they systematically investigated how negative sampling makes sense. However, it is still important to explore how to ensure stable performance in the presence of noise, especially severe noise in datasets.

This paper proposes NRCES (Noise Robust Cross Entropy Sigmoid) loss function for the unlabeled entity problem. Our approach has more flexibility and noise robustness compared to traditional cross entropy loss. Specifically, we employ the contextual embeddings from pre-trained models, and a span-based entity model proposed by Zhong et al.,2021 as the backbone model. To prevent the model from overfitting noise, we believe that the composition of the gradient needs to be rethought, and the gradient contribution of noisy labels needs to be reduced. We change its loss function and conduct experiments on multiple NER datasets. Our method facilitates good learning dynamics at different training stages. We also treat samples differently according to their quality, ensuring they are fully utilized. 

The contributions of this paper are as follows: 
1) We propose a new loss function named NRCES, which is simple to use and has strong noise tolerance by adding a sigmoid term to the cross entropy loss.
2) Our approach achieves a good balance between fast convergence and noise robustness. We combine the training process with the introduced hyperparameter and use a strategy of separately training positive and negative samples.
3) To demonstrate our proposed method's effectiveness and wide applicability, we conduct experiments and analyses on synthetic datasets and real-world datasets, respectively. Our method significantly boosts the system performance under severe noise conditions. Results show that we have significantly improved over previous work and achieved state-of-the-art on real-world datasets.

The rest of the paper is organized as follows. Section 2 reviews approaches previously presented in the literature. Section 3 outlines our model's structure, the shortage of cross entropy loss and our proposed NRCES loss. Section 4 explains the datasets we used and some experiment settings. Section 5 presents experimental results and in-depth analysis. Finally, section 6 concludes the paper and mentions future work.

\section{Related Work}

In this section, we present the overview of distantly supervised NER, the unlabeled entity problem in NER tasks and model learning in noisy environments. 

The mainstream NER systems are designed based on deep neural models, so large annotated corpses with high-quality labels are needed for training. However, those datasets require heavily manual annotation. Some previous methods apply cross-domain learning and semi-supervised learning to solve this problem (Xu et al., 2018; Hao et al., 2021). Several studies use distantly supervised labeled data to alleviate the human burden. Some works treat NER task as the sequence labelling problem and improve performance by modifying CRFs. For example, Fuzzy CRF (Shang et al., 2018) can find high-quality tokens and allows the model to learn potential named entities from them. Partial CRF (Yang et al., 2018; Jie et al., 2019) is a parameter estimation method that marginalizes the likelihood of CRF, making it possible for models to learn from incomplete annotations. Many new training paradigms have been proposed to leverage this challenge. Peng et al. (2019) and Mayhew et al. (2019) treated the distantly supervised NER problem as a PU learning problem. Yang et al. (2018) and Nooralahzadeh et al. (2019) adopted a neural network policy with a reinforcement learning framework to identify noisy instances. Meng et al. (2021) implemented a self-training method with pre-trained language models, an effective technique for improving generalization.

An alternative way to mitigate the annotation cost is to use crowdsourcing, which obtains substantial amounts of labels by hiring crowd workers. The quality of labels collected this way is poor compared to gold-standard annotations labeled by experts. Although we can use label integration to get annotations with high precision by majority voting after sufficient repeated labels are collected, such datasets still suffer from low recall since many entities are missed in complex and ambiguous contexts. 

Li et al.(2020) firstly defined the unlabeled entity problem, where a large number of entities are missed in annotation in NER task. They found that unlabeled entities selected as negative samples lead to significant performance degradation. Based on this finding, they used a span-level cross-entropy loss to eliminate the misguidance of unlabeled entities. Their latest work (Li et al., 2022) rethought the negative sampling mechanism, improved it by weighted sampling distribution, and achieved better loss convergence. However, owing to the limited quality and coverage of the entity in lexicon or knowledge graphs, many unlabeled entities may still not be recalled. Although most methods take measures to mitigate its negative impact, they still seriously affect the model performance.


The performance of deep learning approaches is closely related to the quality of data, and noise labels can severely degrade the performance of models. Therefore many studies have focused on exploring how to learn with noisy labels. Some work redefines new training strategies. For example, Zhang et al. (2021) designed a teacher-student network and co-learning paradigm, which allows the two learning procedures mutually to reinforce together, making full use of the training dataset with one network learning from the other. Another noise learning method is the label correction method which corrects noisy labels to clean labels mainly through the detection of neural network models (Lee et al., 2018; Veit et al., 2017) or with the help of external resources such as knowledge graphs (Li et al., 2017). Apart from the above, loss correction is also one of the existing noise learning methods. One approach is to change the probability of different categories of labels by modelling the noise transition matrix (Pang et al., 2022). Other studies eliminate the noise by defining noise-robust loss functions. Jin et al. (2021) proposed the ER-GCE loss, which achieves good performance on natural language processing datasets compared to previous work.

A distantly supervised NER approach typically deals with two types of label noise (i.e., incomplete and inaccurate annotations). Instead, unlabeled entity problem only focuses on the incomplete one. Whether using distantly supervised labeled data or crowdsourcing, how to maintain the model performance with severe unlabeled entity problem is a question worth investigating. In this study, we mainly focus on how to reduce the impact of the unlabeled false negative samples. Our study is also related to noise-robust loss, where we modify the standard CE loss function by adding an additional term and achieve better robustness.

\section{Method}

In this section, we first introduce the format of samples and the structure of our entity model. Next, under unlabeled entity problem, we analyze the characteristics of CE and its deficiencies, then introduce our loss function NRCES.

\subsection{Entity Model}

A NER model typically takes the text sequence as input and outputs the entities that present in the text and their types. We use a span-based NER model (Li et al., 2021,  Fu et al., 2021, Liu et al., 2022). For input sequence $X=\{x_1,x_2,...x_n\}$, we can enumerate all the spans $S=\{s_1,s_2,...s_m\}$ and assign a label $Y=\{y_1,y_2,...y_m\}$ to each of them for prediction. $(b_i,e_i)$ can be used to denote the start and end index of $s_i$, which corresponds to a string $[x_{b_i},x_{b_{i+1}},...,x_{e_i}]$. For example, for the sentence "Amy left Paris", all the spans can be represented as {(1, 1) , (1, 2), (1, 3), (2, 2), (2, 3),(3, 3)}, and the labels of these spans are all "O" except (1, 1) for Amy, which belongs to "PER",  and (3, 3) belongs to "LOC".

We follow Zhong et al.(2021) and use their entity model as our baseline. First, we use a pre-trained language model (e.g., BERT (Kenton et al., 2019)) to obtain vectorial representations for the input tokens:
\begin{equation}
[h_1,h_2,...,h_n]=BERT(X)
\end{equation}
Next, the span representation can be calculated using the start and end tokens’ representations. A span width feature is also introduced, which can be obtained by a learned embedding lookup table $w$. The span representation $H_{s_i}$ for $s_i$ can be computed as:
\begin{equation}
H_{s_i}=[h_{b_i} ; h_{e_i} ; w_{j-i}]
\end{equation}
where $;$ is column-wise vector concatenation, and $w_{j-i}$ is the $(j-i)$-th embedding of $w$.
Finally, we predict for each span representation by feeding it into a feedforward neural network.
A softmax function is then to get the probabilities over all entity types. Cross Entropy loss can be used when fine-tuning the pre-trained language model. 

\subsection{Learning under Unlabeled Entity Problem}

\subsubsection{Cross Entropy Loss}

When training a span-based NER model, given a span $s_i$ with label $y_i$, the model output is supposed to be $z=[z_1,z_2,...,z_C]$ for all classes where C denotes the number of entity types including non-entity. The Softmax function is usually used in multi-classification problems, where the goal is to optimize the score of the correct category greater than others. The probability for class $k$ is calculated by considering the prediction in all categories mutually exclusive as:
\begin{equation}
p_k=\frac{exp(z_k)}{\sum_{j=1}^C exp(z_j)}, \forall k \in {1,2,...,C}
\end{equation}
The cross-entropy(CE) can be defined as
\begin{equation}L_{ce}=-\hat{y}_i^Tlogf(s_i;\theta)\end{equation}
where the ground truth $\hat{y}_i$ is an one-hot encoded vector, with value being 1 if the index corresponds with the annotated label $y_i$, and 0 otherwise. $f$ is the softmax function. $\theta$ is a set of parameters optimized during training.
The gradient of $L_{ce}$ for $z_k$ is:
\begin{equation}
\frac {\partial L_{ce}}{\partial z_k}=\begin{cases} 
p_k-1,  & \mbox{if $k=y_i$ }\\
p_k,    & \mbox{else }
\end{cases}
\end{equation}

As for the ground-truth label $y_i$, the gradient is $p_{y_i}-1$, and for other $C-1$ classes, the gradients are exactly their corresponding probabilities. This design allows the model to consider the probability distribution under different classes in an integrated manner and ensures the balance of gradient in both positive and negative directions, giving the model better convergence under clean datasets. However, since CE encourages the model to produce logits with larger magnitudes, it leads to the overconfidence issue. For misclassified samples, the network is prone to be more and more confident about its incorrect predictions (Mukhoti et al., 2020). We believe that in the condition of severe noise, careful consideration needs to be given to which components should contribute to the gradient of the loss function. When the model is overfitting on noise, the conflicting noise samples comprise the majority of the loss and dominate the gradient. For an negative sample, when it is classified as an entity and its predicted pliability is close to 1, we want the loss is down-weighted to prevent the misguidance of erroneous samples. We suggest reshaping the loss function to reduce the gradient contributed by those potentially unlabeled entity spans.


\subsubsection{Noise Robust Cross Entropy Sigmoid Loss}

To make the loss function better fit the noisy labels, the CE has been changed in our proposed methods. We add a sigmoid term, which can be defined as
\begin{equation}
L_{sigmoid}=\hat{y}_i^T\sigma(s_i;\theta)
\end{equation}
\begin{equation}
L_{cs}=\beta{L_{ce}}+(1-\beta)L_{sigmoid}
\end{equation}
where $\sigma$ is the sigmoid function. $\beta$ can be computed through
\begin{equation}\beta=exp(-e/w)\end{equation} where $e$ indicates the number of epochs that the current model has been trained, and $w$ is the hyper-parameter to balance two functions.

For the sigmoid function $\sigma(x)=\frac {1}{1+e^{-x}}$, the gradient can be calculated as $\sigma'(x)=\sigma(x)[1-\sigma(x)]$. Compared with CE, which assumes the mutual exclusiveness among classes, sigmoid is more independent and treats every class equally for gradient update. Such characteristic is more compatible with the real-world data, and is more noise-robust than CE. Due to the model's robustness, our model gradually learns to correctly identify the span class as training proceeds, even if the data is noisy. We design an adaptive parameter $w$ to make our approach more effective and flexible. CE is good for convergence which can accelerate learning in the early stage of training, while the sigmoid term can avoid the overfitting issue on noisy data later.

We make further changes to the above loss function to adapt to the unlabeled entity problem. Specifically, we divide the spans into positive samples (entities) and negative samples (i.e., the O label) and update their gradients separately using different strategies:
\begin{equation}
NRCES = L_{ce}\cdot\mathbbm{1} (y_i \in \mathcal{Y}_\text{p}) + L_{cs}\cdot\mathbbm{1} (y_i \in \mathcal{Y}_\text{n})
\end{equation} 
where $\mathcal{Y}_\text{p}$ and  $\mathcal{Y}_\text{n}$ denotes sets of positive and negative labels.
Based on the previous analysis, we believe that the spans labeled as entities usually share higher confidence, so CE is chosen because CE has better convergence. In contrast, due to the unlabeled entity problem, many potential entities are wrongly labelled as non-entities that need to be treated carefully, so we include sigmoid for non-entity spans because of its better noise tolerance. Figure 1 shows how NRCES works.

\begin{figure}
    \centering
    \includegraphics[width=10cm]{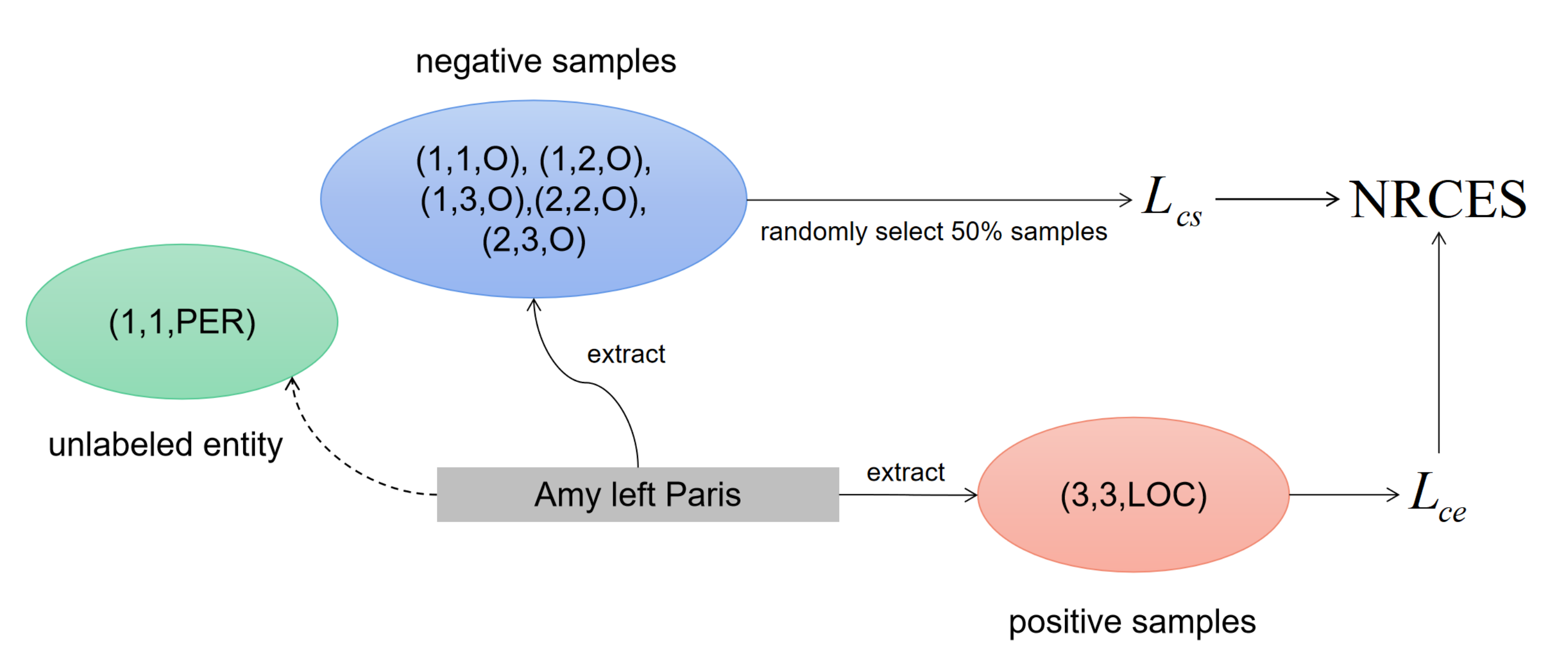}
    \caption{An example to show how NRCES works}
    \label{Figure 1}
\end{figure}

For training, we minimize the loss NRCES. During inference, after enumerating all the spans for the input text, Eq.3 can calculate the prediction probability of the sample under all categories. The label corresponding to the maximum value of the probability is the sample's prediction category.

\section{Experiments}

In this section, we introduce the datasets, evaluation measures and parameter settings of our experiments.

\subsection{Datasets}

Following prior works (Li et al., 2020; 2022), we evaluate NRCES on synthetic and real-world datasets. 

Synthetic datasets are CoNLL-2003 (Sang and De Meulder, 2003) and OntoNotes 5.0 (Pradhan et al., 2013). We follow Ghaddar et al., 2018 and use the same splits for both datasets. We evaluate both in English. The construction of synthetic datasets is based on Li et al., 2020 and will be described in Sec. 5.1. This setting aims at testing our approach in a controlled environment.

Real-world datasets are EC and NEWS, which are from the e-commerce and news domains in Chinese. Yang et al. (2018) collected both of them. They split the data into train, development and test sets and built dictionaries from the training data. Distant supervision was performed on raw data to obtain extra training instances. Due to the insufficient dictionary quality, there is a severe unlabeled entity problem. This setting is to test our method in a more realistic environment.


\subsection{Experiment Settings}

\paragraph{Evaluation Metrics} We follow the standard evaluation protocol and take  F1 score as the evaluation metric. An entity is considered correct only if the boundary and the type are predicted correctly.
\paragraph{Hyper-parameters} 

We adopt pre-trained RoBERTa-base (Liu et al., 2019) as the basic model for English and Google's Chinese BERT-base of Chinese. The batch size is set to 16, the learning rate is 2e-5 for weights in pre-trained LMs, 5e-4 for others. We tune the parameter w from \{2,5,10\}, according to different datasets and noise conditions. We consider spans up to L = 10 words. All entities and half of the randomly selected non-entity span samples are used in training in noise environment. We believe that this is beneficial to the model under unlabeled entity problem.

\section{Results and Analysis}

In this section, we first report the experimental results. Then we investigate the effectiveness of our proposed method by an ablation study, an analysis of the introduced hyper-parameter and a case study.

\subsection{Main Result}

    \begin{table*}
		\centering

		\begin{tabular}{c|ccc}
			\hline
			
			\multirow{2}{*}{Masking Prob.} & \multicolumn{3}{c}{CoNLL-2003}\\
			
			\cline{2-4}
			& Neg.Sampling & Neg.Sampling Variant & NRCES    \\
			
			\hline
			0.5 & $89.22$ & $89.51$ & $\mathbf{89.70}$  \\
			
			0.6 & $87.65$ & $88.03$ & $\mathbf{88.84}$  \\
			
			0.7 & $86.24$ & $86.97$ & $\mathbf{88.73}$  \\
			
			0.8 & $78.84$ & $82.05$ & $\mathbf{87.18}$   \\
			
			0.9 & $51.47$ & $60.57$ & $\mathbf{83.63}$   \\
			\hline
			
		\end{tabular}
		\caption{The comparisons of F1 scores on CoNLL-2003.} 
		\label{tab:Performances on Synthetic Datasets}
	\end{table*}
    \begin{table*}
		\centering
		
		\begin{tabular}{c|ccc}
			\hline
			
			\multirow{2}{*}{Masking Prob.} & \multicolumn{3}{c}{OntoNotes 5.0}\\
			
			\cline{2-4}
			& Neg.Sampling & Neg.Sampling Variant & NRCES    \\
			
			\hline
			0.5 & $88.17$ & $88.31$ & $\mathbf{88.37}$  \\
			
			0.6 & $87.53$ & $88.02$ & $\mathbf{88.30}$  \\
			
			0.7 & $86.42$ & $86.85$ & $\mathbf{87.52}$  \\
			
			0.8 & $85.02$ & $86.12$ & $\mathbf{86.44}$   \\
			
			0.9 & $74.26$ & $80.55$ & $\mathbf{81.90}$   \\
			\hline
			
		\end{tabular}
		\caption{The comparisons of F1 scores on OntoNotes 5.0.} 
		\label{tab:Performances on Synthetic Datasets}
	\end{table*}

\subsubsection{Results on Synthetic Datasets}

    Table 1 and 2 presents the results of our proposed method compared with previous works (Li et al.,2020,2022). We randomly mask some entities as non-entities and see how NRCES reacts to this change. The masking probabilities vary from 0.5 to 0.9. The F1 scores of baselines are copied from Li et al. (2022). Obviously, our method achieves the best performance, especially in high masking probabilities. For example, when masking probability is 0.9, our method improves the F1 score with an increase of 23.06\% on CoNLL-2003 and 1.35\% on OntoNotes 5.0, respectively. It shows that our method is highly robust even if the entities are severely missing. When masking probability changes from 0.5 to 0.9, our performance on CoNLL-2003 drops only 6.07\%, and 6.47\% on OntoNotes 5.0, while Li et al.(2022) drops by 28.94\% on CoNLL-2003 and 7.76\% on OntoNotes 5.0. 
    
	\begin{table*}
		\centering
		
		\begin{tabular}{c|cc}
			
			\hline		
			Method & CoNLL-2003 & OntoNotes 5.0 \\
			
			\hline
			Flair Embedding & $93.09$ & $89.30$ \\
			
			HCR w/ BERT & $93.37$ & $90.30$ \\
			
			BERT-MRC & $93.04$ & $91.11$ \\
			
			BERT-Biaffine Model & $93.5$ & $\mathbf{91.30}$ \\ \hdashline
			
			
			Neg.Sampling & $93.42$ & $90.59$ \\
			Neg.Sampling Variant & $\mathbf{93.68}$ & $91.17$ \\
			\hline
			NRCES & $91.74$ & $89.28$ \\
			
			\hline
		\end{tabular}
		\caption{The experiment results on well-annotated datasets.}
		\label{tab:Performances on Well-annotated Datasets}
	\end{table*}
    
	For a fair comparison, we conducted experiments on well-annotated datasets. Table 3 presents the results. Since our method is specifically designed for the unlabeled entity problem, we are not as good as some classical supervised NER methods, mainly because the model does not predict entities reliably enough (i.e., recall is higher than precision). We recommend using CE directly when the dataset is of high quality to obtain faster convergence. Of course, we can also discard entity spans with confidence below a specific threshold during decoding, allowing the model to obtain a consistent increase in the F1 score.
	
    \begin{table*}
		\centering
		
		\begin{tabular}{c|cc}
			
			\hline
			Method & EC & NEWS \\
			
			\hline 
			
			BERT-MRC & $55.72$ & $74.55$ \\
			
			BERT-Biaffine Model & $55.99$ & $74.57$ \\
			
			\hdashline
			Partial CRF & $60.08$ & $78.38$ \\
			
			Positive-unlabeled (PU) Learning & $61.22$ & $77.98$ \\
			
			Weighted Partial CRF & $61.75$ & $78.64$ \\
			
			Neg.Sampling & $66.17$ & $85.39$ \\

			Neg.Sampling Variant & $67.03$ & $86.15$ \\
			
			\hdashline
			NRCES & $\mathbf{70.66}$ & $\mathbf{89.99}$ \\

			\hline
		\end{tabular}
		\caption{The experiment results on two real-world datasets.}
		\label{tab:Performances on Real-world Datasets}
	\end{table*}

\subsubsection{Results on Real-world Datasets}
	Due to the uneven quality of knowledge resources, many missing entities exist in real-world datasets. Among baselines methods, Partial CRF and PU Learning are distantly supervised methods with noise tolerance. BERT-MRC and Biaffine are classical supervised methods. Both of them show good performance on well-annotated datasets. According to Table 4, we can see that the performance of the supervised methods decreases significantly on real-world datasets. In contrast, NRCES achieves the state-of-the-art on both datasets compared to the previous works, which validates the effectiveness of our method.

    \begin{table*}
        \centering
        
        \begin{tabular}{c|cc}
        \hline
		{ }& NEWS & CoNLL-2003 \\
		\hline
        NRCES  & $91.30(0.70)$  & $90.40(\mathbf{0.17})$  \\
        \hline
            w/o random sampling      & $\mathbf{91.84(0.05)}$  & $89.38(0.60)$         \\
        w/o sigmoid  & $85.15(1.29)$  & $51.88(4.67)$        \\
        w/o separate training  & $86.28(1.84)$  & $44.91(2.25)$         \\
        w/o $\mathbbm{1} (y_i \in \mathcal{Y}_\text{n})$  & $91.00(0.49)$  & $\mathbf{90.54}(0.30)$         \\
        w/o $\mathbbm{1} (y_i \in \mathcal{Y}_\text{p})$  & $84.97(0.23)$  & $38.15(3.90)$         \\
        
        \hline
        \end{tabular}
        \caption{Mean and standard deviation (std.) F1 on dev set.}
        \label{Ablation Study}
    \end{table*}

\subsection{Ablation Study}

    We performed ablation experiments to investigate the effect of each component in our approach: 
    (1) do not perform random sampling, which means training with all negative samples (w/o random sampling); 
    (2) replace the sigmoid term, which means training with CE (w/o sigmoid); 
    (3) do not perform different training strategies with positive and negative samples, which means training all samples with $L_{cs}$ (w/o separate training);
    (4) remove the indicator function for negative samples, which means training with $L_{ce}\cdot\mathbbm{1} (y_i \in \mathcal{Y}_\text{p}) + L_{cs}$. (w/o $\mathbbm{1} (y_i \in \mathcal{Y}_\text{n})$);
    (5) remove the indicator function for entity spans, which means training with $L_{ce} + L_{cs}\cdot\mathbbm{1} (y_i \in \mathcal{Y}_\text{n}$ (w/o $\mathbbm{1} (y_i \in \mathcal{Y}_\text{p})$). Table 5 shows the results on NEWS and CoNLL-2003. We run the model with different seeds and report the mean and standard deviation F1. To simulate a scenario where unlabeled entity problem is extremely severe, we set the masking probability on CoNLL-2003 to 0.8. We demonstrate learning curves with different methods in Figure2.
    
    Interestingly, we found that w/o random sampling on NEWS did not harm the model. We suggest that perhaps sufficient negative samples could sometimes help the model learn comprehensively. It can be seen that the model can have better adaptability by randomly dropping some negative samples before training, especially when the noise is severe. This reduces the number of false negative samples in training. So the impact of missing annotated entities has also been decreased. We also find that even with a large number of entities in the negative samples, our loss function still maintains good performance, i.e. performance on CoNLL-2003 drops only 1.02\% with all negative samples, compared with other ablation experimental results. This proves that the effectiveness comes mainly from our method, not sampling.

    We investigate the effectiveness of the sigmoid term. Theoretically, it can mitigate the impact of those negative samples. It is also our primary noise reduction source. (w/o $\mathbbm{1} (y_i \in \mathcal{Y}_\text{n})$) basically remains the same as our method, which means that the improvement of sigmoid for positive samples is fairly small. Since CE itself already has provided enough high convergence, applying sigmoid to the positive instances may not necessary. Noticing that w/o separate training reduces performance when unlabeled entity problem is severe. We believe the model suffers from underfitting when training with $L_{cs}$. We also tried training the model with more epochs, but the result was not significantly better. We think it seems inappropriate to use $L_{cs}$ alone at a later stage, because it is not enough to provide sufficient loss convergence. Training with only CE (w/o sigmoid) makes the model overfit with noisy labels, leading F1 score decreasing visibly. When using our method in severe noise conditions, we speculate that in the middle and later stages of training, both the augment of CE in positive samples and the tolerance of sigmoid in unlabeled negative samples help the model gradually correct the misguidance.

    Using CE with a large number of false negative samples may bring great harm to model performance. Compared with our method, the F1 value of w/o $\mathbbm{1} (y_i \in \mathcal{Y}_\text{p})$ decreases sharply. It can be seen that the severe misguidance of false negative samples needs to be emphatically avoided in the training progress. Li et al. (2020) also mentioned that treating unlabeled entities as negative instances might lead to worse performance compared with the reduction of annotated entities. The possible reason for the different results between $L_{cs}$ and w/o $\mathbbm{1} (y_i \in \mathcal{Y}_\text{p})$ might also lie in the different levels of misguidance in CE. Since $L_{cs}$ obtained a smaller weight of CE by introduced parameter $w$ during training. 
    
    \begin{figure}
        \centering
        \includegraphics[width=8cm]{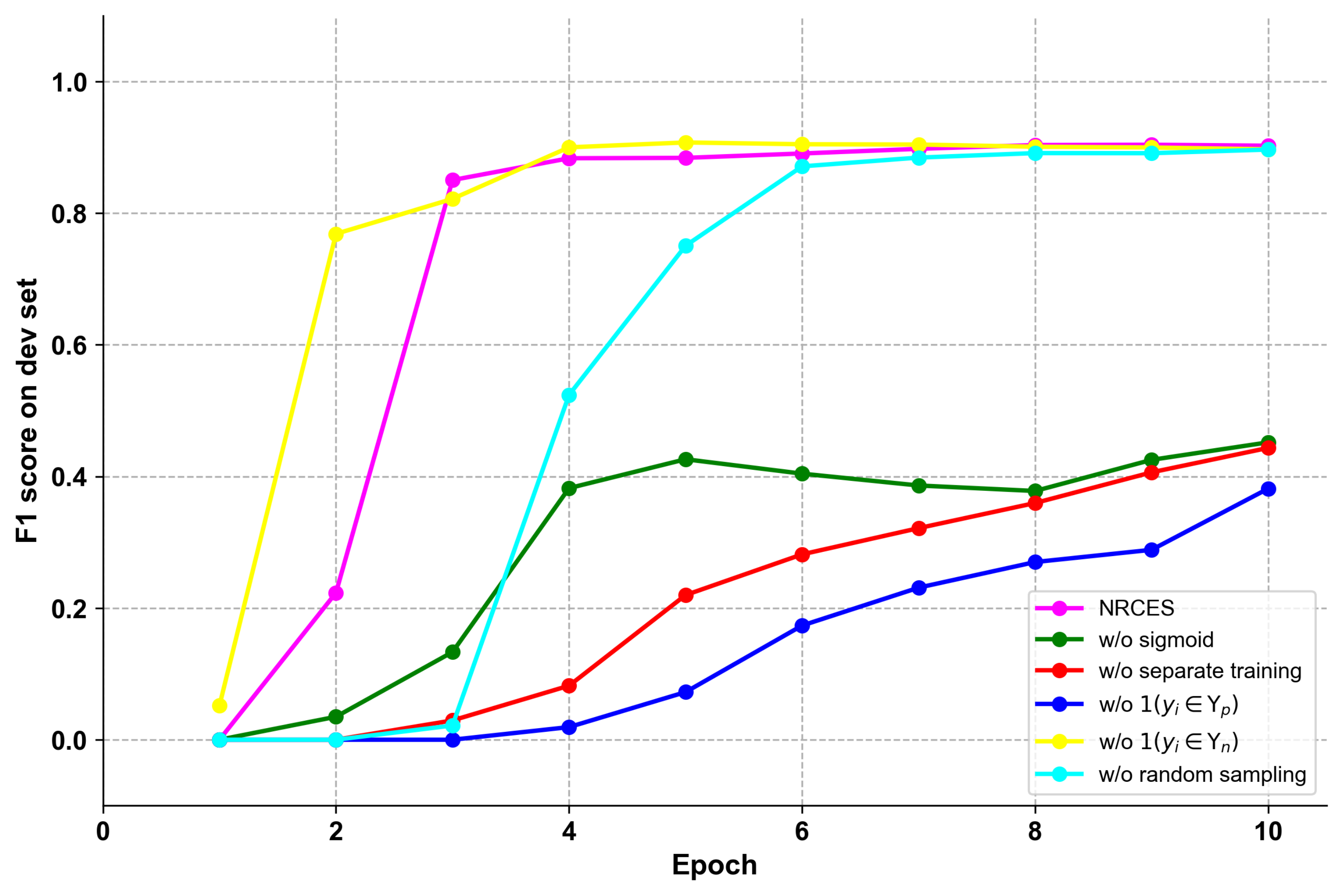}
        \caption{F1 on CoNLL03 with 0.8 masking prob.}
        \label{Figure 1}
    \end{figure}

    It is worth noting that model stability on different methods varies greatly. Since CE’s strong convergence to both positive and negative samples, the performance of models training with CE (w/o sigmoid) depends on the possible best performance they can achieve before being overfitted with noise. However, this also makes the training unstable due to the extremely high conflict contributed by unlabeled entities. Similarly, we find that w/o separate training and w/o $\mathbbm{1} (y_i \in \mathcal{Y}_\text{p})$ also show poor stability due to their deficiencies. We believe that stability is an important metric for evaluating performance. Our method has good stability because we have a relatively high and stable recall, which also illustrates the superiority of our method.

    \begin{figure}
        \centering
        \includegraphics[width=8cm]{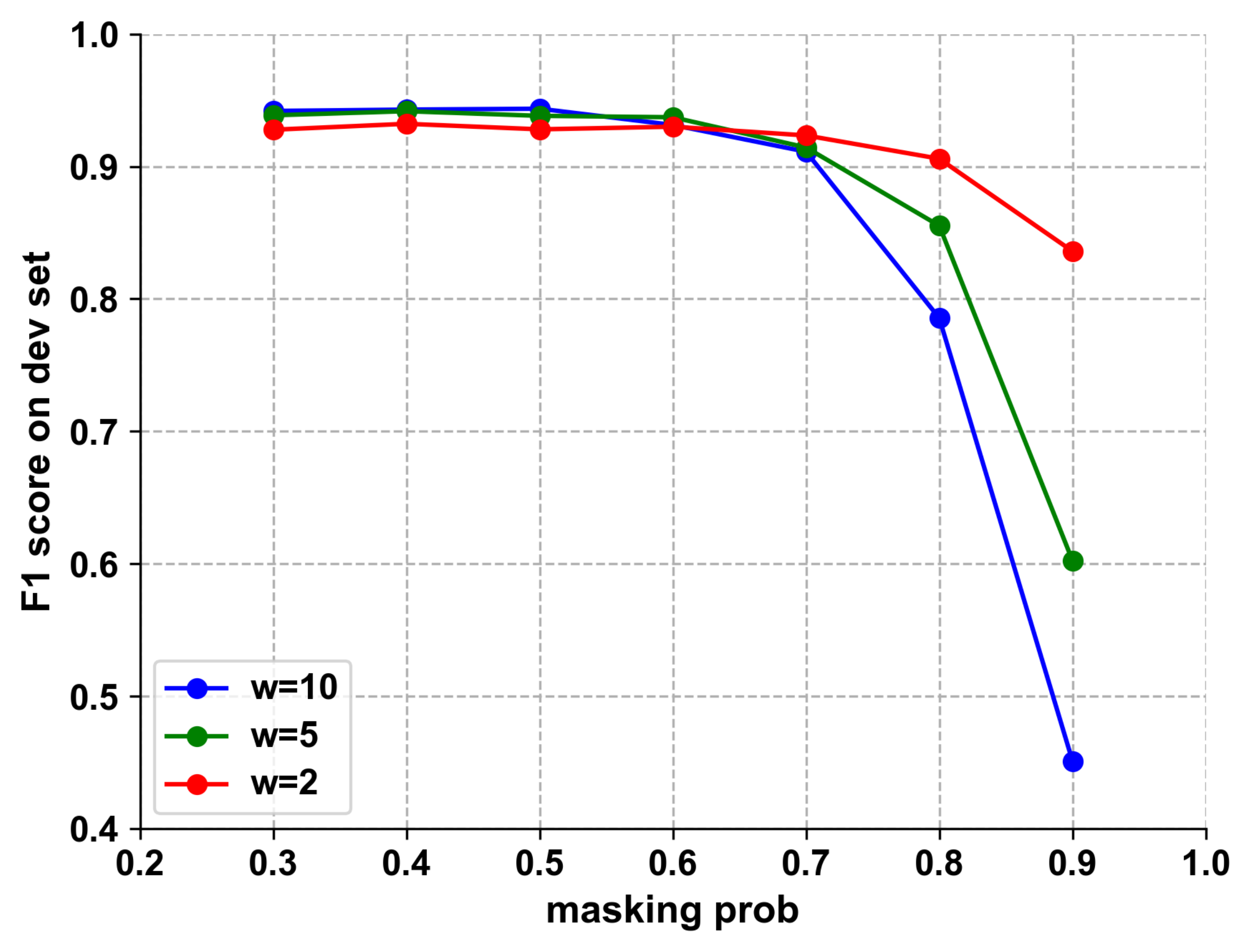}
        \caption{F1 on CoNLL03 with different masking prob.}
        \label{Figure 1}
    \end{figure}

\subsection{Sensitivity of the introduced hyper-parameter}
    
    We study the influence of $w$ in Eq.8. Intuitively, $w$ balances the convergence and noise-robustness contributed by CE and sigmoid, respectively. The optimal $w$ should also vary on different datasets and noise conditions. When $w$ → $1$, the loss function approximates CE with high convergence, and when $w$ → $+\infty$, the loss function remains CE on positive samples and approximates $L_{sigmoid}$ on negative samples with high noise robustness. To simulate different noisy environments, we train the model by varying the value of $w$ range in \{2, 5, 10\} with masking probability from 0.3 to 0.9. Figure 3 shows the F1 score on CoNLL-2003. 
    
    The performance is rather insensitive to $w$ when the masking probability is less than 0.7. When the noise is weak, a large $w$ helps the model converge fast and learn from hard samples quickly. A small $w$, however, makes model performance drop a bit. This is probably because too many spans are predicted as entities when w is small. Hard samples are not sufficiently trained, leading to underfitting.

    $w$ shows more sensitivity when the masking probability is greater than 0.7. As the noise is more severe, a smaller $w$ allows the sigmoid to better help the model discard the gradient contributed by the missing annotated entities, achieving better performance. A larger $w$ reduces performance more significantly, showing the harm of model overfitting noise.

    Overall, as the severity of unlabeled entity problem increases, a small $w$ shows better stability. Although it may not be the best in a weak noise condition, it maintains good performance in the extremely severe case due to sigmoid’s strong tolerance.
    
\newcommand{\per}[1]{\textcolor{red}{[#1]$_{\text{PER}}$}}
\newcommand{\org}[1]{\textcolor{blue}{[#1]$_{\text{ORG}}$}}
\newcommand{\loc}[1]{\textcolor{purple}{[#1]$_{\text{LOC}}$}}

\subsection{Case Study}

    \begin{table*}[t]
    \centering
    \scalebox{0.8}{
    \begin{tabular}{l}
    \toprule
    \textbf{Ground Truth}: \org{Leicestershire} beat \org{Somerset} by an innings and 39 runs. \\
    \textbf{NRCES}: \org{Leicestershire} beat \org{Somerset} by an innings and 39 runs. \\
    \textbf{CE Loss}: \org{Leicestershire} beat Somerset by an innings and 39 runs.  \\
    \midrule
    \textbf{Ground Truth}: The world 's costliest footballer \per{Alan Shearer} was  named as the new \\\loc{England} captain on Friday . \\
    \textbf{Might be Incorrectly Predicted as}: The world 's costliest footballer \per{Alan Shearer} was \\ named as the \loc{new England} captain on Friday . \\
    \textbf{NRCES}: The world 's costliest footballer \per{Alan Shearer} was  named as the new \\ \loc{England} captain on Friday . \\
    \textbf{CE Loss}: The world 's costliest footballer \per{Alan Shearer} was  named as the new England \\ captain on Friday . \\
    \bottomrule
    \end{tabular}
    }
    \caption{
    Case study with diffenrent sentences.
    }
    \label{tab:case}
    \end{table*}

    Finally, we perform case studies on our method. We select two examples in the development set of CoNLL-2003, exploring how models are learned during the training process. We use the dataset with a mask probability of 0.8. Experiments are conducted using CE and our loss function.
    
    \begin{figure}[htbp]
        \centering
        \begin{minipage}[t]{0.48\textwidth}
        \centering
        \includegraphics[width=5cm]{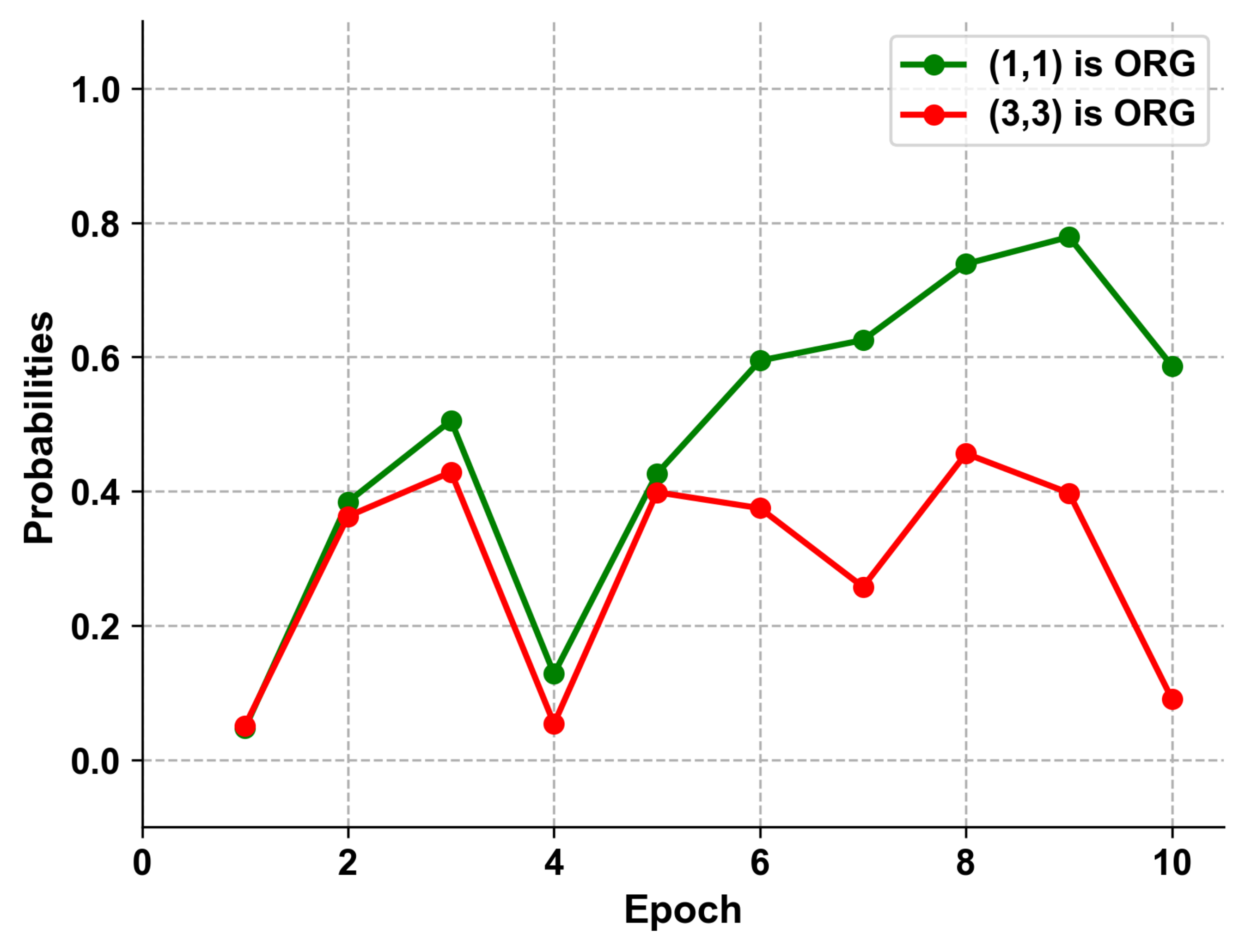}
        \caption{Sentence 1 with CE}
        \end{minipage}
        \begin{minipage}[t]{0.48\textwidth}
        \centering
        \includegraphics[width=5cm]{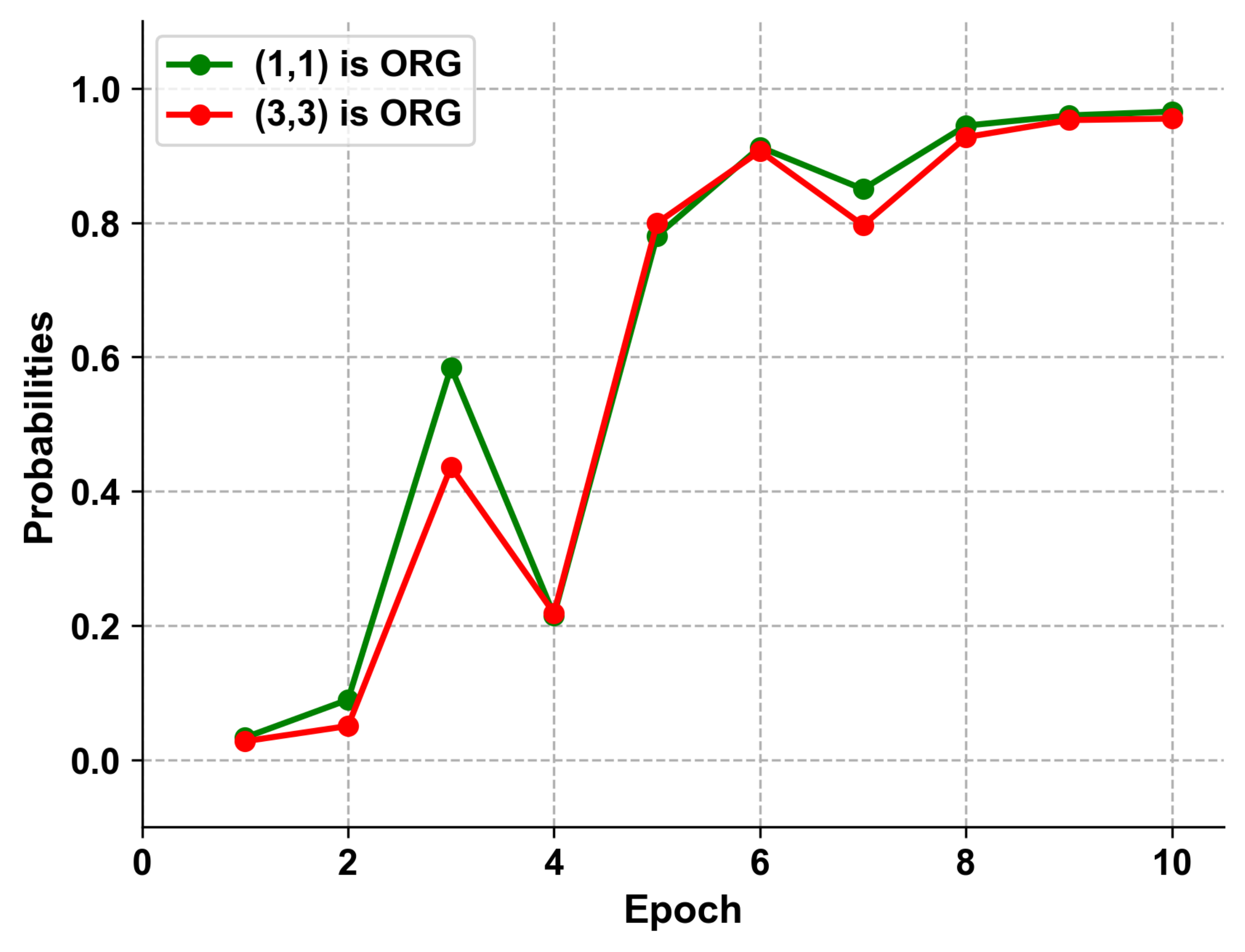}
        \caption{Sentence 1 with NRCES}
        \end{minipage}
    \end{figure}
    \begin{figure}[htbp]
        \centering
        \begin{minipage}[t]{0.48\textwidth}
        \centering
        \includegraphics[width=5cm]{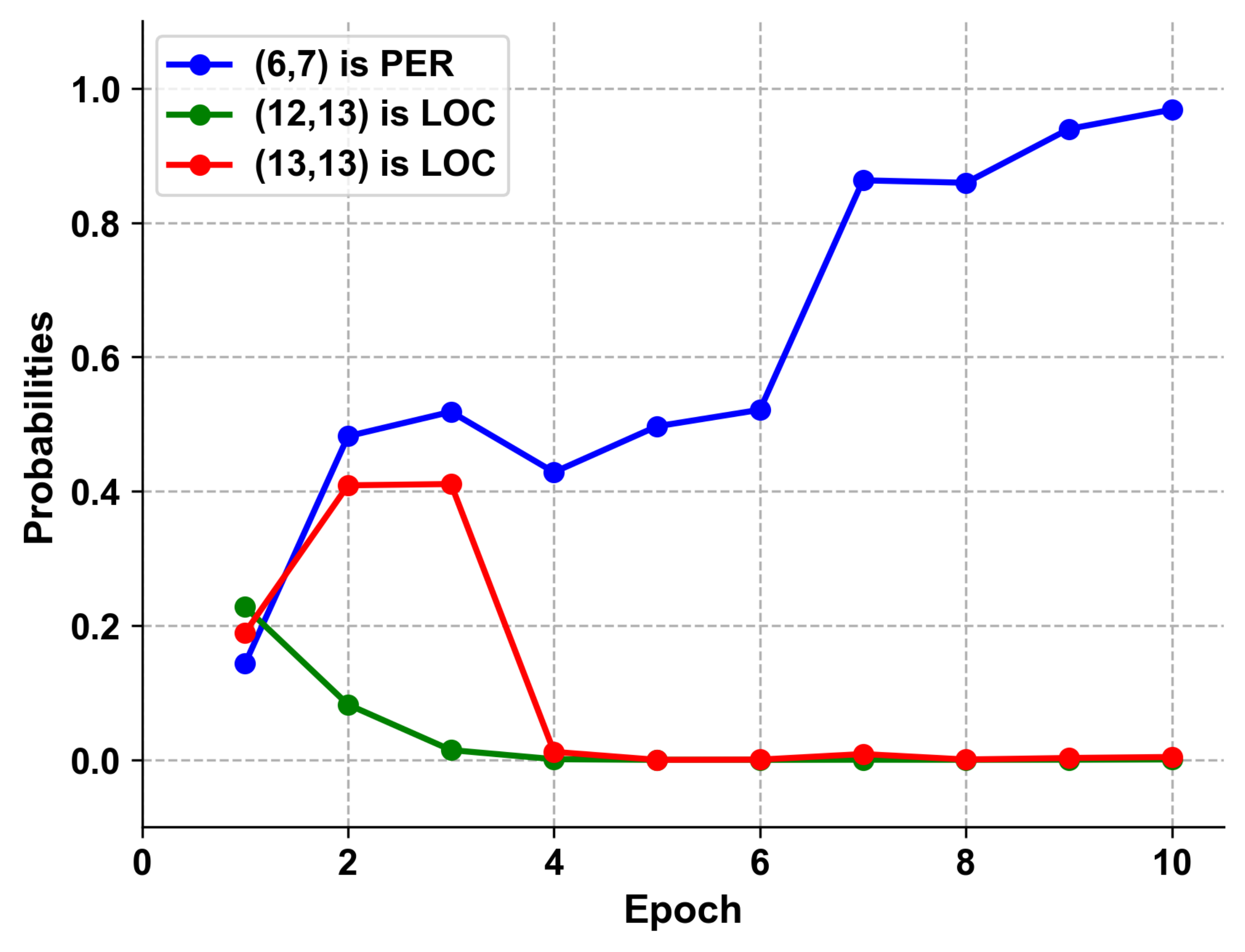}
        \caption{Sentence 2 with CE}
        \end{minipage}
        \begin{minipage}[t]{0.48\textwidth}
        \centering
        \includegraphics[width=5cm]{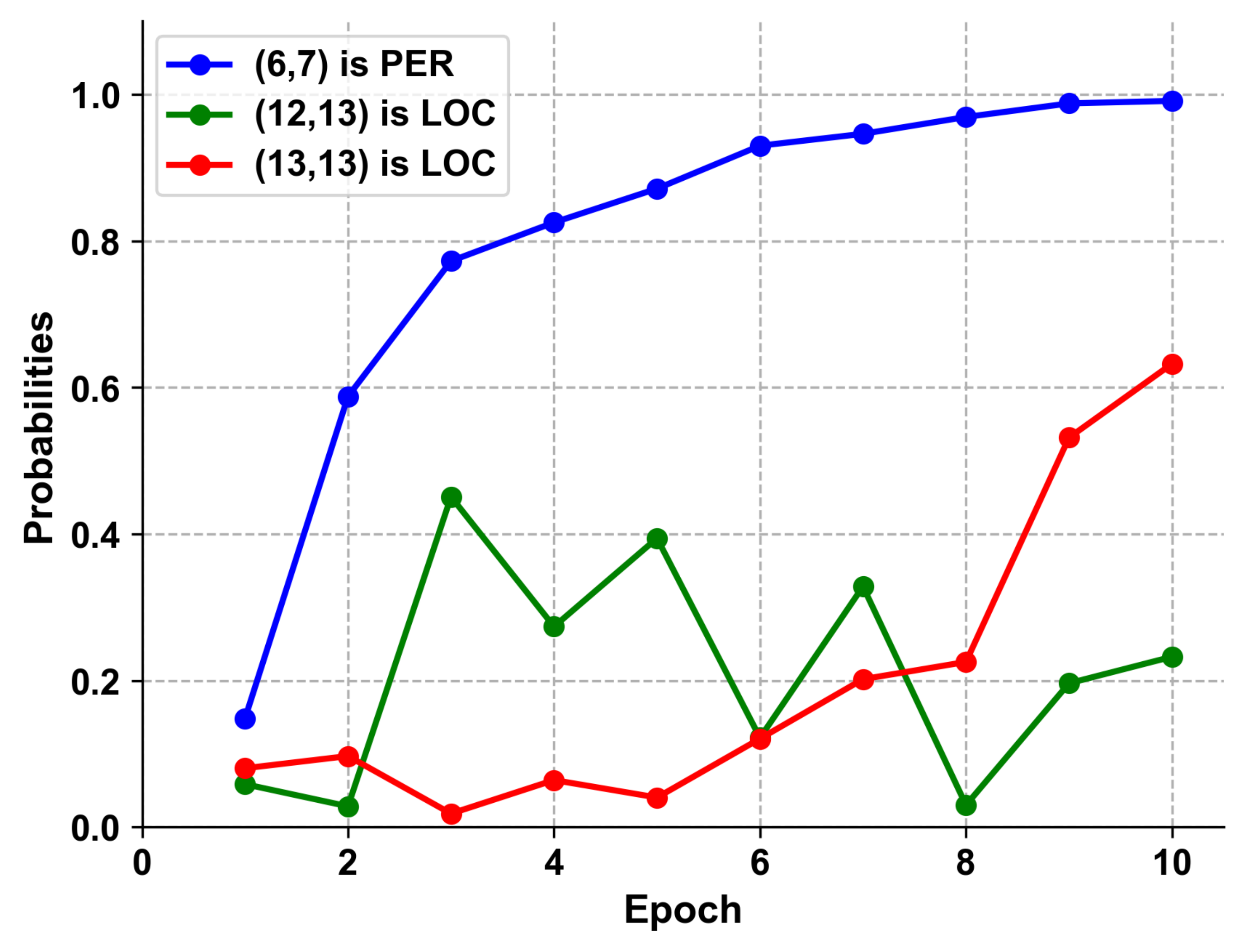}
        \caption{Sentence 2 with NRCES}
        \end{minipage}
    \end{figure}

    Given a sentence, we record the predicted probability of a span to be an entity. For sentence 1, "Leicestershire beat Somerset by an innings and 39 runs." Figures 4 and 5 record the change in probability for "Leicestershire" and "Somerset", respectively. As training proceeds, the model trained by the CE overfits the mislabeled data, resulting in entities being extracted with difficulty. Our loss function shows better noise tolerance, so the model can consistently extract entities later in training.

    We further explore how the model performs on hard samples. For the NER task, boundary detection is more ambiguous and error-prone than entity type classification. For example, for sentence 2, "The world 's costliest footballer Alan Shearer was named as the new England captain on Friday", where the boundary detection of "England" and "new England" need to be considered in context. Figures 6 and 7 record the predicted probabilities changes for three potential entity spans "Alan Shearer", "new England", and "England". For the easy sample "Alan Shearer", the performance of our loss function remains more stable as the training continues. For the hard sample "England", our loss function has better extraction capability and generalizes better for more accurate entity detection.

\section{Conclusion and Future Work}

In this paper, a loss function named NRCES is proposed to deal with the unlabeled entity problem in named entity recognition. It is simple to use and achieves a good balance between convergence and robustness. Experimental results on synthetic and real-world datasets demonstrate its effectiveness, especially under severe noise conditions. Notably, we achieve the state-of-the-art on two real-world datasets. Furthermore, we study the impact of our method in-depth through other auxiliary experiments. 

For future work, since there are both hard samples and unlabeled entities in the negative samples, how to maximize the utilization of hard samples and avoid the negative impact of unlabeled entities as much as possible can be explored. We can also probe other approaches to balance convergence and noise tolerance capability. Moreover, the loss function can be applied to other fields of denoising tasks.

\end{document}